# Robust Bayesian optimization with reinforcement learned acquisition functions

Zijing Liu [a,c,d], Xiyao Qu [a,c], Xuejun Liu [a,c,*], Hongqiang Lyu [b,*]

[a] Ministry of Industry and Information Technology Key Laboratory of Pattern Analysis and Machine Intelligence, College of Computer Science and Technology, Nanjing University of Aeronautics and Astronautics, Nanjing 211106, China

[b] College of Aerospace Engineering, Nanjing University of Aeronautics and Astronautics, Nanjing 210016, China

[c] Collaborative Innovation Center of Novel Software Technology and Industrialization, Nanjing 210023, China

[d] Key Laboratory of Aerodynamic Noise Control and State Key Laboratory of Aerodynamics, Mianyang 621000, China

## Abstract

In Bayesian optimization (BO) for expensive black-box optimization tasks, acquisition function (AF) guides sequential sampling and plays a pivotal role for efficient convergence to better optima. Prevailing AFs usually rely on artificial experiences in terms of preferences for exploration or exploitation, which runs a risk of a computational waste or traps in local optima and resultant re-optimization. To address the crux, the idea of data-driven AF selection is proposed, and the sequential AF selection task is further formalized as a Markov decision process (MDP) and resort to powerful reinforcement learning (RL) technologies. Appropriate selection policy for AFs is learned from superior BO trajectories to balance between exploration and exploitation in real time, which is called reinforcement-learning-assisted Bayesian optimization (RLABO). Competitive and robust BO evaluations on five benchmark problems demonstrate RL's recognition of the implicit AF selection pattern and imply the proposal's potential practicality for intelligent AF selection as well as efficient optimization in expensive black-box problems.

---

∗ Corresponding authors.
  *E-mail addresses*: zijing.liu@nuaa.edu.cn (Z. Liu), quxiyao@nuaa.edu.cn (X. Qu), xuejun.liu@nuaa.edu.cn (X. Liu), hongqiang.lu@nuaa.edu.cn (H. Lyu).

# 1. Introduction

Bayesian optimization (BO) [1] is an effective optimization framework for expensive black-box problems by virtue of Bayesian theory, where Gaussian process (GP) is generally used as a prior belief over the objective function and sequentially refined using optimal observations measured by an acquisition function (denoted by AF, also called infill sampling criterion) for purpose of accurate localization of the promising region. However, prevailing AFs are either endowed with changeless preferences for exploration or exploitation due to their fixed formulas, or imposed with pre-set adjustment rules, which may not always be appropriate during the optimization and possibly cause either a computational waste or traps in local optima. It brings us questions: is it possible to achieve a fast and optimal convergence via real-time AF adjustment and is there some inherent pattern in the adjustment?

Herein, we propose a BO framework with learned sequential AF selection by means of reinforcement learning (RL) [2] technology, called reinforcement-learning-assisted Bayesian optimization (RLABO). In the method, appropriate selection policy for AFs at varied optimization stages is learned from superior BO trajectories, so as to contribute to efficient optimization in a promising direction. Intuitively, AF selection only depends on the current posterior GP, the optimal candidate measured by the incumbent AF in turn updates the posterior GP, and the AF to be selected in the next interaction is accordingly adjusted by the AF selection policy. During the optimization, the objectives of the candidates evaluate the incumbent AFs. Therefore, AF selection for the whole optimization process can be formalized as a Markov decision process (MDP), whose solution is in reinforcement learning context.

Our main contributions are summarized in the following three folds:

● We propose the idea of data-driven AF selection to improve BO performance.

● Specifically, we model the sequential AF selection in a BO process as an MDP and resort to powerful RL technologies.

● Evaluation on five benchmark problems shows the competitive and robust BO performance armed with the learned AF selection, implying the ability of the proposed RLABO framework for a good recognition of the implicit AF selection pattern.

## 2. Related work

A lot of researches on BO have been devoted to a good trade-off between exploration and exploitation. They can be basically classified into multiple-point infilling and single-point infilling. The former obtain multiple candidates in each iteration usually through independent optimization [3] or multi-objective optimization [4,5] for multiple AFs with diverse preferences for exploration or exploitation. Robust performance is achieved for taking both exploration and exploitation into consideration at the same time, but the number of function evaluations conclusively becomes multifold regardless of parallel computing.

This work falls into the latter category. It is quite popular to obtain trade-offs between exploration and exploitation by introducing controllable parameters, such as the improvement margin $\xi$ in probability of improvement (PI) [6]/expected improvement (EI) [7,8], the $g$-order moment of the improvement in generalized expected improvement (GEI) [9], the parameter $t$ weighting the improvement moments of different orders in moment-generating-function-based AF (MGF-based AF) [10], as well as the weighting parameter $\beta$ between prediction and uncertainty in upper confidence bound (UCB) [11,12] or the weights in linearly integrated AFs. All these parameters need to be carefully chosen to ensure competence for different problems. Yet, the changeless preference may cause slow localization of global optimum directions in early stages and insufficient search accuracy for optima in later stages when facing tough tasks with spiked optima in large search spaces. To address the issue, two mainlines of researches on adaptive AFs have been developed. One is to adjust these trade-off parameters as the optimization progresses. For instance, a schedule for $\xi$ decreasing from a high value to zero is recommended for adaptive PI, a similar "cooling" schedule using a look-up table is proposed to apply on $g/t$ in GEI/MGF-based AF [10,13], and the trade-off parameters $\beta_t$ in two variants of UCB, Gaussian process upper confidence bound (GP-UCB) [14] and randomized Gaussian process upper confidence bound (RGP-UCB) [15], are functions of iteration $t$ and naturally adapt to the optimization processes. Unfortunately, most of the adaption rules still rely on experience and inappropriate ones may be counterproductive. Another kind is to adopt portfolios of AFs governed by certain strategies, such as entropy search portfolio (ESP) [16] selecting individual AF at each iteration according to an entropy search (ES) [17] meta-criterion, and GP-Hedge and its variants [18,19,20]/adaptive cost-aware Bayesian optimization (ACBO) [21] according to predefined rewards. They are of good performance by fully using the available information. Differently, this work realizes adaptive AF selection in a data-driven way which utilizes entire trajectory information instead of the currently available. Besides, it will respectively fall into the two mainlines by using an infinite set of candidate AFs with a same formulation and a continuous trade-off parameter or a limited number of ones.

## 3. Reinforcement-learning-assisted Bayesian optimization

In general RL, an actor learns to perform actions from continuous interactions with the environment to achieve a maximal cumulative reward. In the AF selection learning task, a neural network (NN) serves as the AF selection actor while a BO framework is perceived as the environment. Figure 1 visualizes the interactions between them. The NN actor takes GP information as a state input, and gives a sampled AF from the predicted action probability distribution as an action output. According to the selected AF, a new observation is obtained in BO environment for GP refinement and its function value is mapped to a feedback reward to the NN actor for action policy updates. The specific design of the involved essential ingredients is as follows:

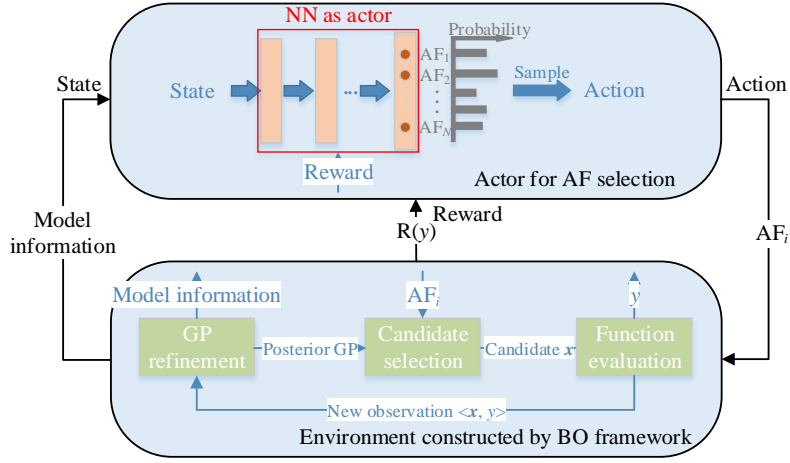

**Fig. 1.** Learning AF selection using RL framework.

1) **State:** We encode the length scale of the employed Matern32 kernel as well as the number and the dimensional variances of the observations into the state vector $s_t$. The former influences the fluctuation characteristics of the predicted function, which is intuitively related to the preference for exploration or exploitation of the AF to be selected; the latter two measure the distribution characteristics of the observations and reflect the reliability of GP prediction to some extent. The combination provides a sufficient basis for AF selection decision.

2) **Action:** Upper confidence bound (UCB) [12] is a group of simple and effective AFs that combines the prediction and uncertainty in a linear way. By adjusting the weight parameter, the tradeoff between exploration and exploitation can be easily achieved. Therefore, UCBs with different weights 0, $2.576^{i-1}$ ($i$=1,2,3), $\infty$ are used herein as candidate actions $a_t$. Certainly the weight can also be set within a certain range for sake of continuous actions and a resultant general applicability.

3) **Reward:** According to BO's goal of efficient optimization, reward $r_t$ is devised as the improvement rate of the currently observed $y_t$ relative to the incumbent $\max_{i\in\{1,\ldots,t-1\}} y_i$. It will be 0 if there is no improvement. The reward is formulated as

$$r_t = \max\left(y_t - \max_{i\in\{1,\ldots,t-1\}} y_i, 0\right) \Big/ \left[\ln(t+1)+1\right], \quad (1)$$

where the denominator is designed to avoid excessive punishment for later improvements.

4) **State transition:** The transition probability from the current state to the next after taking an action is formed

by the uncertainties involved in BO environment.

5) **Policy update:** We use proximal policy optimization (PPO) [22] in actor-critic style to update the action policy, where both the actor and the critic are multilayer perceptrons (MLP) with two hidden layers of 64 units and tanh nonlinearities.

The loss function for policy update herein is in a weighted sum form as

$$L(\boldsymbol{\theta}) = L_{\text{CLIP}}(\boldsymbol{\theta}) + w_1 L_{\text{SE}}(\boldsymbol{\theta}) + w_2 L_{\text{entropy}}(\boldsymbol{\theta}), \tag{2}$$

where $\boldsymbol{\theta}$ is the splicing parameter vector of the actor and the critic, and $w_1$ and $w_2$ are the weights. The first term $L_{\text{CLIP}}$ denotes the clipped loss, and it is formulated as

$$L_{\text{CLIP}}(\boldsymbol{\theta}) = -\sum_{<s_t,a_t>} \min\left[\rho_t(\boldsymbol{\theta})\hat{A}_t, \text{clip}(\rho_t(\boldsymbol{\theta}), 1-\varepsilon, 1+\varepsilon)\hat{A}_t\right], \tag{3}$$

where $\varepsilon$ is the clipping parameter and $\rho_t(\boldsymbol{\theta})$ is the probability ratio $p_{\boldsymbol{\theta}}(a_t|s_t)/p_{\boldsymbol{\theta}_{\text{old}}}(a_t|s_t)$ during update. $\hat{A}_t$ is the advantage estimator of the actor performance over the critic evaluation and defined as

$$\hat{A}_t = R_t - V_{\boldsymbol{\theta}}(s_t), \tag{4}$$

where $R_t = \sum_{t'=t}^{T} \gamma^{t'-t} r_{t'}$ is the cumulative discounted reward with discount parameter $\gamma$ from the current timestep $t$ to the end of the episode and $V_{\boldsymbol{\theta}}(s_t)$ is the evaluated state value given by the critic. The second term $L_{\text{SE}}$ denotes the squared error loss between the actor performance and the critic evaluation as

$$L_{\text{SE}}(\boldsymbol{\theta}) = \sum_{<s_t,a_t>} \left(R_t - V_{\boldsymbol{\theta}}(s_t)\right)^2. \tag{5}$$

The third term $L_{\text{entropy}}$ denotes the entropy loss of the policy $\pi_{\boldsymbol{\theta}}$ which is used to ensure exploration during training and formulated as

$$L_{\text{entropy}}(\boldsymbol{\theta}) = -\sum_{<s_t,a_t>} \text{H}\left(\pi_{\boldsymbol{\theta}}(s_t)\right). \tag{6}$$

The proposed RLABO method contains two phases: AF selection policy training and BO execution, which are respectively given in Algorithms 1 and 2.

**Algorithm 1** RLABO training phase.

**Input:** Policy update number $M$; episode number $N$ used in each policy update; timestep number $T$ in one episode; epoch number $K$ in each policy update;

**Output:** Trained policy $\pi_{\theta*}$;

1: Initialize parameter vector $\theta_{\text{old}}$;
2: **for** $m$=1, 2, …, $M$ **do**
3:     **for** $n$=1, 2, …, $N$ **do**
4:         Initialize GP and organize into state $s_{n1}$;
5:         **for** $t$=1, 2, …, $T$ **do**
6:             Run policy $\pi_{\theta_{\text{old}}}$ with current state $s_{nt}$ and sample $a_{nt} = \text{AF}_i$;
7:             Optimize $\text{AF}_i(\boldsymbol{x})$ to obtain a candidate $\boldsymbol{x}_{nt}$;
8:             Evaluate objective function value $y_{nt}$ of $\boldsymbol{x}_{nt}$ and compute reward $r_{nt}$ as Eq. (1);
9:             Update GP using $<\boldsymbol{x}_{nt}, y_{nt}>$ and organize into a new state;
10:        **end for**
11:    **end for**
12:    Optimize Eq. (2) w.r.t. $\theta$ for $K$ epochs using the $NT$ $<s_{nt}, a_{nt}, r_{nt}>$ to obtain $\theta^*$;
13:    $\theta_{\text{old}} \leftarrow \theta^*$;
14: **end for**
15: **return** $\theta^*$

---

**Algorithm 2** RLABO execution phase.

**Input:** Trained policy $\pi_{\theta*}$;

**Output:** Optimized objective function value $y^*$;

1: Initialize GP and organize into state $s_1$;
2: **for** $t$=1, 2, …, $T$ **do**
3:     Run policy $\pi_{\theta*}$ with current state $s_t$ and elect $a_t = \text{AF}_i$ with maximal probability;
4:     Optimize $\text{AF}_i(\boldsymbol{x})$ to obtain a candidate $\boldsymbol{x}_t$;
5:     Evaluate objective function value $y_t$ of $\boldsymbol{x}_t$;
6:     Update GP using $<\boldsymbol{x}_t, y_t>$ and organize into a new state;
7: **end for**
8: **return** $y^* \leftarrow \max\{y_1, \ldots, y_T\}$

## 4. Evaluation

The proposed RLABO is evaluated on five benchmark problems, Ackley, Levy, Griewank, Schwefel and Eggholder [23], and their function contours are shown in Fig. 2. Divergent numbers of local minimizers can be seen among them, which correspond to divergent difficulties for optimization. We respectively perform episodes of training to basic convergences shown in Fig. 3. The learned actors are used for AF selection in BO tests and the optimization processes are compared to those with fixed candidate AFs. Figure 4 shows the comparative results for BO tests on the five benchmarks. Note that the benchmarks are in negative to form maximization problems in the experiment.

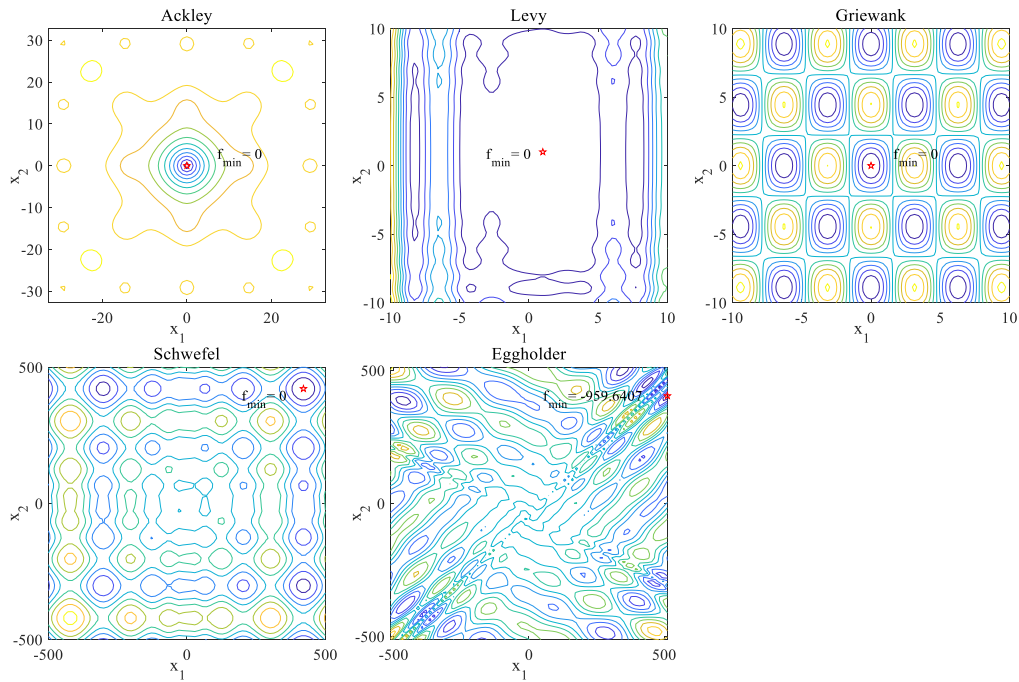

**Fig. 2.** Contours of benchmark problems. Dark contours indicate small function values, among which the closed ones encircled by light contours indicate the local minima. The denser the contours, the larger the function gradients. The red stars represent the global minimizers.

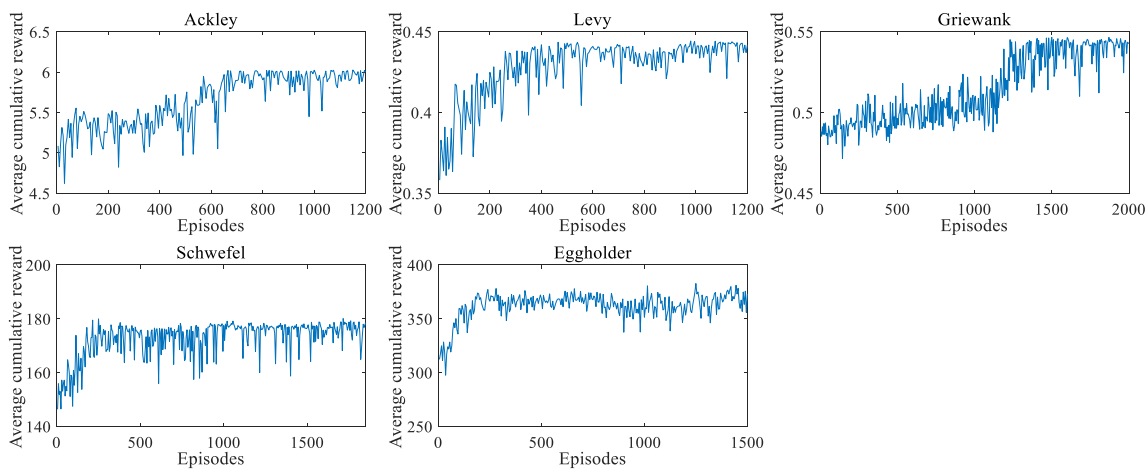

**Fig. 3.** Learning curves on benchmarks. Averages of the cumulative rewards in every five episodes are shown. General convergences are achieved in the training processes for the benchmarks.

From Fig. 4 we can observe that the BO processes with the learned AF selection policies achieve the most efficient convergences to better/competitive optimized results on all benchmarks and no one else is always superior. Among the competitors, the most explorative exhibits the best performance in complex problems such as Eggholder and Schwefel but performs worst in simple problems such as Ackley, Levy and Griewank, while it is almost the opposite for the most exploitative one. The other three almost invariably achieve modest results for their trade-off between exploration and exploitation. This underlines the fact that no fixed AF is universally applicable to various kinds of problems which calls for self-adaptive AFs.

The proposed RLABO with data-driven AF selection exhibits steady superiority over the competitors in all the considered benchmarks as expected. Furthermore, selection of AFs in RLABO for these problems shows a general pattern that explorative AFs tend to be selected more frequently for a general construction of GP as the complexity of the problem increases and preferences for exploration or exploitation are tentatively switched according to the current GP posteriori and sampling. RLABO learns the pattern under the instruction of the devised reward function rather than exactly copying the best trajectories, resulting in faster and better convergences than the competitors in most of the benchmarks.

To sum up, the proposal is effective and robust and the AF selection policy is intuitive and intelligent, which reflects that RL catches the internal pattern in AF adaption.

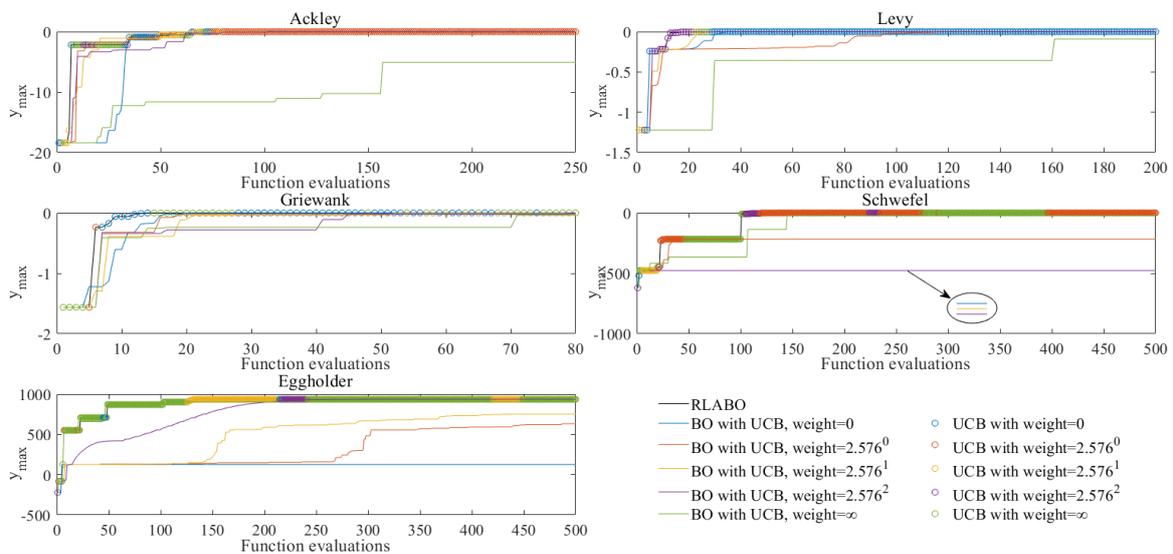

**Fig. 4.** Comparative BO processes on benchmarks. The lines in different colors are the optimization processes

obtained by different methods, and the different colors on the RLABO process distinguish the AFs selected by the trained policy. Note the arrow shows the methods coincident in the optimization processes.

## 5. Conclusions

We propose to learn sequential AF selection with real-time adjustment to steer clear of the dilemma caused by intrinsic sampling preferences in common AFs in a BO framework. We further reveal the essence of the task as an MDP which can be easily solved with RL technologies and devise the key elements in this context. Competitive and robust results demonstrate the effectiveness of our proposal and imply its potential practicality for intelligent AF selection as well as efficient optimization in expensive black-box problems.

As preliminary validation for the proposal, the problems used for AF selection policy training are used for testing and its generalization is not studied in this work. Yet it is worth noting that the proposal learns to select sequential AFs instead of sequential candidates and the devised state does not directly contain sample information. The purpose is to learn a general and adaptive BO framework to automatically adjust its preference for exploration or exploitation according to the specific problems and their optimization stages. This helps to achieve generalization on unseen optimization problems and thus provides an effective way to solve expensive problems through training on similar non-expensive ones. We are working on the validation and further improvement of the proposal's generalization.

**Declaration of competing interest**

The authors declare that they have no known competing financial interests or personal relationships that could have appeared to influence the work reported in this paper.

**Acknowledgements**

This work was supported by the Key Laboratory of Aerodynamic Noise Control [grant number ANCL20190103]; the State Key Laboratory of Aerodynamics [grant number SKLA20180102]; and the Aeronautical Science Foundation of China [grant numbers 2018ZA52002, 2019ZA052011].

**References**

[1] B. Shahriari, K. Swersky, Z.Y. Wang, R.P. Adams, N.D. Freitas, Taking the human out of the loop: a review of Bayesian optimization, in: Proceedings of the IEEE, 2015, pp. 148–175. https://doi.org/10.1109/JPROC.2015.2494218.

[2] R.S. Sutton, A.G. Barto, Reinforcement learning: an introduction, IEEE Transactions on Neural Networks 9 (5) (1998). https://doi.org/10.1109/TNN.1998.712192.

[3] F. Hutter, H.H. Hoos, K. Leyton-Brown, Parallel algorithm configuration, in: Proceedings of the 6th International Conference on Learning and Intelligent Optimization, 2012, pp. 55–70. https://doi.org/10.1007/978-3-642-34413-8_5.

[4] B. Bischl, S. Wessing, N. Bauer, K. Friedrichs, C. Weihs, Moimbo: Multiobjective infill for parallel model-based optimization, in: International Conference on Learning and Intelligent Optimization. Springer, 2014, pp. 173–186. https://doi.org/10.1007/978-3-319-09584-4_17.

[5] H. Wang, M. Emmerich, T. Back, Balancing risk and expected gain in kriging-based global optimization, in: 2016 IEEE Congress on Evolutionary Computation (CEC), IEEE, 2016, pp. 719–727. https://doi.org/10.1109/CEC.2016.7743863.

[6] H.J. Kushner, A new method of locating the maximum of an arbitrary multipeak curve in the presence of noise, Journal of Basic Engineering 86 (1) (1964) 97-106. https://doi.org/10.1115/1.3653121.

[7] J. Mockus, V. Tiesis, A. Zilinskas, The application of Bayesian methods for seeking the extremum, in: Toward Global Optimization, Elsevier, 1978, pp. 117-128.

[8] D. Lizotte, Practical Bayesian optimization, PhD thesis, University of Alberta, Edmonton, Alberta, Canada, 2008.

[9] M. Schonlau, W.J. Welch, D.R. Jones, Global versus local search in constrained optimization of computer models, Lecture Notes-Monograph Series, 1998, pp. 11-25.

[10] H. Wang, B.V. Stein, M. Emmerich, T. Back, A new acquisition function for Bayesian optimization based on the moment-generating function, 2017 IEEE International Conference on Systems, Man and Cybernetics (SMC), IEEE, 2017, pp. 507-512. https://doi.org/10.1109/SMC.2017.8122656.

[11] D.D. Cox, S. John, A statistical method for global optimization, in: Proc. IEEE Conference on Systems, Man and Cybernetics (SMC), IEEE, 1992, pp. 1241-1246. https://doi.org/10.1109/ICSMC.1992.271617.

[12] P. Auer, Using confidence bounds for exploitation-exploration tradeoffs, Journal of Machine Learning Research 3 (11) (2002). https://doi.org/10.1162/153244303321897663.

[13] M.J. Sasena, P. Papalambros, P. Goovaerts, Exploration of metamodeling sampling criteria for constrained global optimization, Engineering optimization 34 (3) (2002) 263-278. https://doi.org/10.1080/03052150211751.

[14] N. Srinivas, A. Krause, S.M. Kakade, M. Seeger, Gaussian process optimization in the bandit setting: no regret and experimental design, in: Proceedings of the 27th International Conference on Machine Learning, 2010, pp. 1015–1022. https://doi.org/10.48550/arXiv.0912.3995.

[15] J. Berk, S. Gupta, S. Rana, S. Venkatesh, Randomised Gaussian process upper confidence bound for Bayesian optimization, International Joint Conference on Artificial Intelligence, 2020, pp. 2284-2290. https://doi.org/10.24963/ijcai.2020/312.

[16] B. Shahriari, Z. Wang, M.W. Hoffman, A.B. Cote, N.D. Freitas, An entropy search portfolio for Bayesian optimization, in: Proc. NIPS workshop Bayesian optim, 2014.

[17] P. Hennig, C. Schuler, Entropy search for information-efficient global optimization, The Journal of Machine Learning Research 13 (6) (2012) 1809-1837.

[18] M. Hoffman, E. Brochu, N.D. Freitas, Portfolio allocation for Bayesian optimization, Conference on Uncertainty in Artificial Intelligence, 2011, pp. 327-336.

[19] T.d.P. Vasconcelos, D.A.R.M.A. de Souza, C.L.C. Mattos, J.P.P. Gomes, No-PASt-BO: Normalized portfolio allocation strategy for Bayesian optimization, 2019 IEEE 31st International Conference on Tools with Artificial Intelligence (ICTAI), 2019, pp. 561-568. https://doi.org/10.1109/ICTAI.2019.00084.

[20] T.d.P. Vasconcelos, D.A.R.M.A. de Souza, G.C.d.M. Virgolino, C.L.C. Mattos, J.P.P. Gomes, Self-tuning portfolio-based Bayesian optimization, Expert Systems With Applications 188 (2022) 115847. https://doi.org/10.1016/j.eswa.2021.115847.

[21] P. Luong, D. Nguyen, S. Gupta, S. Rana, S. Venkatesh, Adaptive cost-aware Bayesian optimization, Knowledge-Based Systems 232 (1) (2021) 107481. https://doi.org/10.1016/j.knosys.2021.107481.

[22] J. Schulman, F. Wolski, P. Dhariwal, A. Radford, O. Klimov, Proximal policy optimization algorithms, 2017, arXiv preprint

arXiv:1707.06347v2.

[ 23 ] Virtual library of simulation experiments: test functions and datasets - optimization test problems. https://www.sfu.ca/~ssurjano/optimization.html, 2022 (accessed 27 August 2022).